\documentclass{article}
\usepackage{PRIMEarxiv}
\usepackage[utf8]{inputenc}
\usepackage[T1]{fontenc}
\usepackage{url}
\usepackage{booktabs}
\usepackage{amsfonts}
\usepackage{nicefrac}
\usepackage{microtype}
\usepackage{lipsum}
\usepackage{fancyhdr}
\usepackage{graphicx}
\graphicspath{{figures/}{media/}}
\usepackage{float}
\usepackage{placeins}
\usepackage{caption}
\usepackage{multirow}
\usepackage{diagbox}
\usepackage[affil-it]{authblk}
\usepackage[numbers,sort&compress]{natbib}

\usepackage[colorlinks=true, citecolor=blue, linkcolor=blue, urlcolor=cyan]{hyperref}

\title{Ran Score: a LLM-based Evaluation Score for Radiology Report Generation}

\author[1]{Ran Zhang}
\author[1,2]{Yucong Lin$^{*}$}
\author[3]{Zhaoli Su}
\author[3]{Bowen Liu}
\author[1]{Danni Ai}
\author[1]{Tianyu Fu}
\author[1]{Jingfan Fan}
\author[1]{Yuanyuan Wang}
\author[1]{Mingwei Gao}
\author[8,9]{Yuwan Hu}
\author[10]{Shuya Gao}
\author[5,6]{Jingtao Li}
\author[1,2]{Deqiang Xiao}
\author[4]{Hong Song$^{*}$}
\author[7]{Hongliang Sun$^{*}$}
\author[1]{Jian Yang$^{*}$}
\affil[1]{School of Optics and Photonics, Beijing Institute of Technology, Beijing 100081, China}
\affil[2]{Zhengzhou Research Institute, Beijing Institute of Technology, Zhengzhou 450003, China}
\affil[3]{School of Medical Technology, Beijing Institute of Technology, Beijing 100081, China}
\affil[4]{School of Computer Science and Technology, Beijing Institute of Technology, Beijing 100081, China}
\affil[5]{Department of Gastroenterology, China-Japan Friendship Hospital, Beijing 100029, China}
\affil[6]{NHC Key Laboratory of Clinical Big Data Standardization \& Integration, Beijing 100029, China}
\affil[7]{Department of Radiology, China-Japan Friendship Hospital, Beijing 100029, China}
\affil[8]{Department of Radiology, China-Japan Friendship Hospital (Institute of Clinical Medical Sciences), Beijing 100029, China}
\affil[9]{Chinese Academy of Medical Science \& Peking Union Medical College, Beijing 100730, China}
\affil[10]{Department of Radiology, Peking University China-Japan Friendship School of Clinical Medicine, Beijing 100029, China}

\begin{document}
\maketitle
\vspace{-45pt}
\begin{center}
   *Co-corresponding authors\\
   Yucong Lin: linyucong@bit.edu.cn,\ Hong Song: songhong@bit.edu.cn\\
   Hongliang Sun: stentorsun@gmail.com,\ Jian Yang: jyang@bit.edu.cn
\end{center}
\vspace{10pt}
\vspace{12pt}
\begin{abstract}
Chest X-ray report generation and automated evaluation are limited by poor recognition of low-prevalence abnormalities and inadequate handling of clinically important language, including negation and ambiguity. We develop a clinician-guided framework combining human expertise and large language models for multi-label finding extraction from free-text chest X-ray reports and use it to define Ran Score, a finding-level metric for report evaluation. Using three non-overlapping MIMIC-CXR-EN cohorts from a public chest X-ray dataset and an independent ChestX-CN validation cohort, we optimize prompts, establish radiologist-derived reference labels and evaluate report generation models. The optimized framework improves the macro-averaged score from 0.753 to 0.956 on the MIMIC-CXR-EN development cohort, exceeds the CheXbert benchmark by 15.7 percentage points on directly comparable labels, and shows robust generalization on the ChestX-CN validation cohort. Here we show that clinician-guided prompt optimization improves agreement with a radiologist-derived reference standard and that Ran Score enables finding-level evaluation of report fidelity, particularly for low-prevalence abnormalities.
\end{abstract}

% keywords can be removed

\section{Introduction}
Artificial intelligence (AI)-generated radiology reports hold promise for reducing diagnostic variability and improving workflow efficiency\cite{white2022,gundogdu2021}. However, their limited clinical adoption stems largely from a disconnect between conventional technical metrics---such as lexical overlap measures (e.g., BLEU\cite{papineni2002}, ROUGE\cite{lin2004}, CIDEr\cite{vedantam2015})---and true clinical validity. Radiologists typically show strong agreement on clear-cut diagnoses (for example, Pneumothorax), but exhibit considerable inter-reader variability in ambiguous cases, such as pulmonary nodules\cite{huh2023}. Existing evaluation frameworks often fail to adequately assess diagnostic reasoning consistency, negation handling, or precise entity matching, and recent work has highlighted the continuing need for more clinically meaningful and standardized evaluation paradigms\cite{delbrouck2025,rexrank2024,xu2024}. This limitation erodes confidence in AI-generated outputs.

Domain-specific metrics, such as CheXbert F1\cite{smit2020} and RadGraph F1\cite{jain2021}, provide greater clinical relevance than surface-level similarity scores (Kendall's $\tau > 0.6$) by prioritizing accurate entity extraction. Nevertheless, recent NLP advances highlight ongoing limitations: BLEU scores show only weak correlation with radiologist error judgments ($\tau = 0.462$)\cite{yu2023}, whereas entity-aware metrics such as RaTEScore achieve stronger alignment ($\tau = 0.653$) through improved negation handling and synonym normalization\cite{zhao2024}. Even state-of-the-art models like GPT-4\cite{adams2023} frequently miss critical findings in the absence of explicit clinical context, underscoring the benefit of Human--LLM collaboration\cite{sun2023}. Recent multimodal report generation studies have likewise shown continued performance limitations despite substantial architectural advances\cite{jiang2025,zhang2024bionlp,srivastav2024}. Open-source LLMs have recently shown performance comparable to CheXbert\cite{smit2020} for chest report labeling, yet they still struggle with rare abnormalities\cite{dorfner2024}. Privacy-preserving open-source models, including Vicuna, additionally enable secure, local deployment for radiology report labeling\cite{mukherjee2023}.

Despite the scale of MIMIC-CXR\cite{johnson2019}, high-quality annotations for low-prevalence abnormalities remain scarce, and existing labels such as CheXbert\cite{smit2020} are often limited by automated extraction errors in rare or negation-heavy cases. Many existing chest X-ray report generation studies based on MIMIC-CXR rely on CheXbert-derived labels for auxiliary supervision, multi-task learning, or automated evaluation. However, the CheXbert label space is limited in scope and does not adequately capture several clinically relevant or low-prevalence findings, which may weaken both supervision during model development and assessment of report fidelity at evaluation. To address this gap, we develop a clinician-guided Human--LLM collaborative framework that achieves high agreement with the radiologist-derived reference standard for multi-label extraction (macro-averaged F1 score = 0.956). Based on this framework, we construct and publicly release the first large-scale, clinically aligned multi-label annotation resource covering the full MIMIC-CXR corpus (\textasciitilde377{,}000 reports with 21 standardized abnormality categories). This resource provides broader and more clinically meaningful label coverage for both chest X-ray report generation and radiology-oriented language-model evaluation. Compared with the more limited CheXbert label space, our 21-category annotation scheme additionally captures findings such as tracheobronchial abnormalities, cavity and cyst, mediastinal abnormalities, calcification, and pulmonary vascular abnormalities.

We hypothesized that integrating dynamic radiologist feedback into large language model prompting would improve agreement with a radiologist-derived reference standard for multi-label finding extraction from chest X-ray reports and outperform existing automated labelers and surface-form evaluation metrics. We therefore developed and validated a clinician-guided Human--LLM collaborative framework for automated multi-label finding extraction across a standardized 21-label chest X-ray taxonomy.

The Human--LLM collaborative framework offers a generalizable human-in-the-loop approach for aligning large language models with expert domain knowledge in high-stakes clinical settings. In contrast to conventional fine-tuning or retrieval-augmented methods\cite{jeong2024}, which often require substantial computational resources and large annotated datasets, our error-driven prompt refinement strategy enables transparent optimization without retraining the underlying model. By incorporating clinicians' diagnostic reasoning, the framework addresses common failure modes, including synonym variation, negation detection, and class imbalance\cite{bu2024,yin2025}. This makes the framework especially suitable for resource-limited environments and privacy-sensitive applications using open-weight models. The framework may also be extensible to other medical text-processing tasks in which diagnostic accuracy and interpretability are essential.

Together, these contributions establish a clinician-guided framework for clinically aligned finding extraction from chest X-ray reports and define Ran Score as a finding-level evaluation metric for radiology report generation. By coupling radiologist-derived reference labels with transparent prompt optimization, this work also provides a reproducible resource for evaluating and improving report fidelity across MIMIC-CXR-EN and ChestX-CN clinical settings. We further release the code, optimized prompts and complete multi-label finding annotations for the full MIMIC-CXR corpus to support reproducibility and future development in radiology artificial intelligence.

\section{Datasets}

This retrospective analysis used data from two sources: the
MIMIC-CXR-EN dataset \cite{johnson2019} derived from MIMIC-CXR (version
2.0.0), and the independent ChestX-CN dataset. MIMIC-CXR is a publicly
available dataset comprising 377,110 chest X-ray images in DICOM format
from 227,835 radiographic studies involving 65,379 patients treated at
the Beth Israel Deaconess Medical Center Emergency Department, Boston,
MA, USA, between 2011 and 2016. The dataset includes free-text
radiology reports and has been de-identified in accordance with the US
Health Insurance Portability and Accountability Act of 1996 (HIPAA)
Safe Harbor requirements, with protected health information removed.

\subsection{MIMIC-CXR Dataset}
From MIMIC-CXR, three non-overlapping report cohorts were constructed.
First, 3,000 chest X-ray reports with complete \textit{Findings} and
\textit{Impression} sections were randomly sampled for exploratory
entity extraction and construction of the standardized 21-label
taxonomy. Second, 300 additional reports were assigned to a
MIMIC-CXR-EN development cohort for clinician-guided prompt
optimization and radiologist reference-standard construction. Third, an
independent MIMIC-CXR-EN test cohort of 3,000 reports was reserved for
downstream radiology report generation evaluation. No overlap was
allowed among the three cohorts. Non-chest reports were excluded.

\subsection{ChestX-CN Dataset}
The ChestX-CN dataset comprised 150 anonymized chest X-ray reports
collected from a tertiary academic medical center and its affiliated
community hospitals in China. Reports were selected to capture
variation across participating hospitals and reporting styles. All
reports underwent standardized de-identification, including (1) regular
expression-based removal of protected health information, such as names
and exact dates; (2) manual verification by hospital information
department staff; and (3) replacement of original identifiers with
anonymous surrogate codes. The dataset structure was aligned with that
of the MIMIC-CXR report set to support consistent comparative analyses.

\section{Methods}

\subsection{Reference Label Construction}
To define the standardized finding label set, we randomly sampled 3,000
chest X-ray reports from MIMIC-CXR that did not overlap with either the
MIMIC-CXR-EN development cohort or the MIMIC-CXR-EN test cohort.
Disease-related entities were extracted from these reports and grouped
by anatomical location and radiographic semantics to support
construction of a standardized 21-label taxonomy (Fig.~1a). This step
was exploratory and intended only to inform label design. Large
language models were used only to support exploratory text mining and
clustering during taxonomy development; no LLM-generated binary labels
were used in the final reference standard.

\begin{figure}[!t]               
  \centering
  \includegraphics[width=0.85\textwidth]{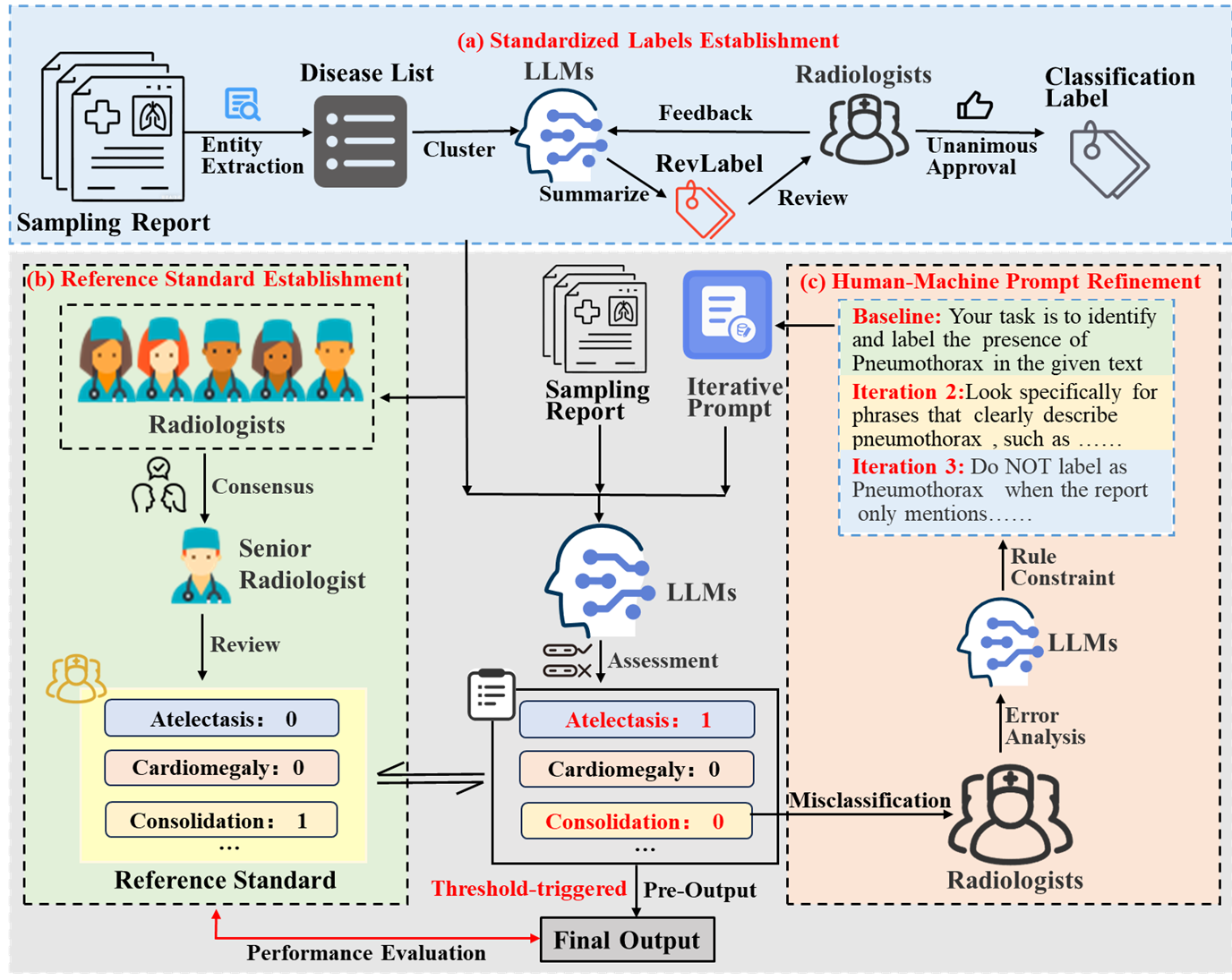}
  % figure file placeholder
  \caption{\textbf{Human--LLM collaborative framework for multi-label finding extraction from chest X-ray reports. (a) Exploratory extraction and clustering of disease-related entities from 3,000 chest X-ray reports to inform the definition of standardized finding labels. (b) Establishment of the diagnostic reference standard through independent binary annotation of reports by six board-certified thoracic radiologists. (c) Illustration of the Human--LLM collaborative workflow using pneumothorax as an example, showing iterative clinician-guided prompt refinement based on error analysis to improve finding-level extraction performance toward higher agreement with the radiologist-derived reference standard.}}
  \label{fig:framework}
\end{figure}

\subsection{Reference Standard Establishment}

Six board-certified thoracic radiologists independently annotated all 300 reports in the MIMIC-CXR-EN development cohort and all 150 reports in the ChestX-CN validation cohort using the predefined 21-label schema (Fig. 1b). Majority vote ($\geq 4/6$) was used to establish the final reference standard for each label. Binary labels were assigned according to predefined criteria: presence (1) required explicit mention of the abnormality or unequivocal diagnostic evidence; absence (0) was assigned only for definitive negation or statements indicating very low likelihood. Uncertain cases (–1) were excluded, consistent with clinical practice where diagnostic uncertainty prompts further investigation or follow-up rather than definitive classification.
All six radiologists acted as fully independent and equal readers, with no adjudicator or hierarchical review. The final reference standard was determined solely by majority vote ($\geq 4$ out of 6 agreement) across the independent assessments. Consequently, the reference standard preserves the inherent inter-reader variability of clinical practice, as it derives from independent annotations without consensus-building,
adjudication, or post-hoc harmonization.

\subsection{Human-LLM Cooperation Framework for Finding Label Extraction}
We developed a Human--LLM collaborative framework for automated multi-label finding extraction from chest X-ray reports, illustrated using Pneumothorax as an example (Fig.~1c). The workflow uses iterative prompt engineering with an initial structured template that enforces strict binary classification, excluding uncertain labels to reflect real-world clinical decision-making.

The framework combines clinician-guided reasoning with LLM processing to generate interpretable binary predictions from unstructured reports. The diagnostic threshold was defined through three rounds of Delphi consensus with six thoracic radiologists as $\geq 90\%$ label-specific accuracy (Cohen's $\kappa \geq 0.90$), representing a prespecified target for high agreement with the radiologist-derived reference standard. This threshold is based on evidence that radiologists show near-perfect agreement (Cohen's $\kappa$ or raw agreement $> 0.95$) for unequivocal chest X-ray findings, but exhibit greater variability in ambiguous cases.

We selected Qwen3-14B as the baseline model because its open-weight availability enabled local deployment, supported reproducible prompt optimization, and provided a stable foundation for subsequent cross-model comparisons. Optimization followed a closed-loop process wherein, when performance fell below predefined clinical thresholds, radiologist-guided error analysis prompted targeted revision of the prompt. Revisions focused on three aspects: (1) incorporating core terms and their synonyms; (2) clarifying vague or ambiguous descriptions; and (3) adding positive and negative examples. This iterative, human-in-the-loop approach systematically addressed failure modes, achieved $\geq 90\%$ label-specific accuracy across all categories, and remained fully transparent and extensible to other abnormalities.

The optimized prompt template was applied to compare Qwen3-14B with Qwen-Plus, GPT-3.5-turbo, GPT-4o-mini and DeepSeek-R1 using API calls (20--25 March 2025). All models produced binary predictions for the 21 standardized labels. Predictions were compared in a blinded manner against the radiologist-derived reference standard. Label-specific accuracy and Clopper--Pearson 95\% confidence intervals were calculated for each model. Key performance metrics (macro-averaged F1 score, precision, recall and accuracy) were compared across the models.

Initial evaluations using the structured prompt template showed persistent performance gaps for low-prevalence abnormalities, mainly attributable to class imbalance in the dataset. To mitigate this, we iteratively refined the prompts on the 300-report development set by selectively adding exemplars for underrepresented conditions. Radiologist-guided error analysis identified specific failure modes. For each underrepresented label, carefully selected exemplars were incorporated to address synonym variation, negation patterns, and contextual ambiguity while maintaining strict binary outputs. Refinements continued over multiple rounds until the prespecified performance threshold was achieved for all targeted labels on the development set.

\subsection{Evaluation of Chest X-ray Report Generation Models}
Generated reports were produced for the independent 3,000-report
MIMIC-CXR-EN test cohort, which did not overlap with either the
3,000-report taxonomy-construction cohort or the 300-report
MIMIC-CXR-EN development cohort. Generated reports were produced by
seven representative chest X-ray report generation models---RGRG,
XrayGPT, R2GenGPT, PromptMRG, LLM-RG4, Libra-1.0-3B, and
MedKit\cite{tanida2023,thawakar2024,wang2023,jin2024,wang2025,zhang2025,su2025}---using
the original chest X-ray images from the held-out MIMIC-CXR studies.

The diagnostic quality of these generated reports was assessed
automatically by applying the optimized and frozen Qwen3-14B-based
extraction framework to both the generated reports and the corresponding
original MIMIC-CXR reference reports. Binary predictions were obtained
for the 21 standardized labels from both sets of reports. Predictions
from the generated reports were then compared with those from the
corresponding original reference reports using macro-averaged and
micro-averaged F1 score, precision, recall, and accuracy. We term the
macro-averaged F1 score derived from this comparison Ran Score. Ran
Score was selected as the primary evaluation metric because it assigns
equal weight to each finding label and is therefore more sensitive to
clinically important omissions in low-prevalence abnormalities.

\subsection{Performance and Agreement Analysis}
Performance metrics---including accuracy, precision, recall, and
macro-averaged and micro-averaged F1 scores---were calculated with
Clopper--Pearson 95\% confidence intervals. Inter-rater agreement among
the six thoracic radiologists was evaluated using Cohen's $\kappa$. 
Differences before and after prompt optimization were assessed with two-sided tests;
P \textless{} 0.05 was considered statistically significant. All
analyses were conducted using Python 3.11.

\section{Results}

\subsection{Dataset characteristics}
The study included 6,450 chest X-ray reports from unique individuals
across two distinct clinical settings: 6,300 reports from the
MIMIC-CXR-EN dataset (median age 58 years {[}IQR 40--72{]}; 56\% male)
and 150 reports from the ChestX-CN dataset (median age 58 years {[}IQR
45--70{]}; 49.3\% male). These datasets capture diverse patient
populations and report complexities, enabling robust evaluation of the
framework's performance and cross-lingual generalization (Table~1,
Fig.~1). Table~1 presents the key demographic and clinical
characteristics, including data collection periods, imaging modality,
median report length, and the most prevalent abnormalities (for
example, lung opacity was present in 22.3\% of MIMIC-CXR-EN reports and
40\% of ChestX-CN reports).

\begin{table}[!htbp]
  \centering
  \caption{\textbf{Demographic and Clinical Characteristics of the Study Cohorts}}
  \label{tab:cohorts}
  \begin{tabular}{lcc}
    \toprule
    Characteristic & MIMIC-CXR-EN (n=6,300) & ChestX-CN (n=150) \\
    \midrule
    Data collection period & 2011--2016 & 2024.11--2025.2 \\
    Age (years), median (IQR) & 58 (40--72) & 58 (45--70) \\
    Male, n (\%) & 3,528 (56) & 74 (49.3) \\
    Imaging modality & Chest X-ray & Chest X-ray \\
    Report length (words), median & 75 (26--202) & 78 (55--110) \\
    \midrule
    \multicolumn{3}{l}{\textit{Common abnormalities}} \\
    Lung Opacity & 22.3\% & 40\% \\
    No Finding & 33\% & 30\% \\
    Consolidation & 4.6\% & 4\% \\
    
    \bottomrule
    \multicolumn{3}{@{}p{\linewidth}@{}}{\footnotesize \raggedright \textit{Note.} --- Values are reported as counts with percentages in parentheses unless otherwise specified. Age is presented as median (IQR). CXR, chest X-ray.} \\
  \end{tabular}
\end{table}

\begin{figure}[!t]               
  \centering
  \includegraphics[width=0.85\textwidth]{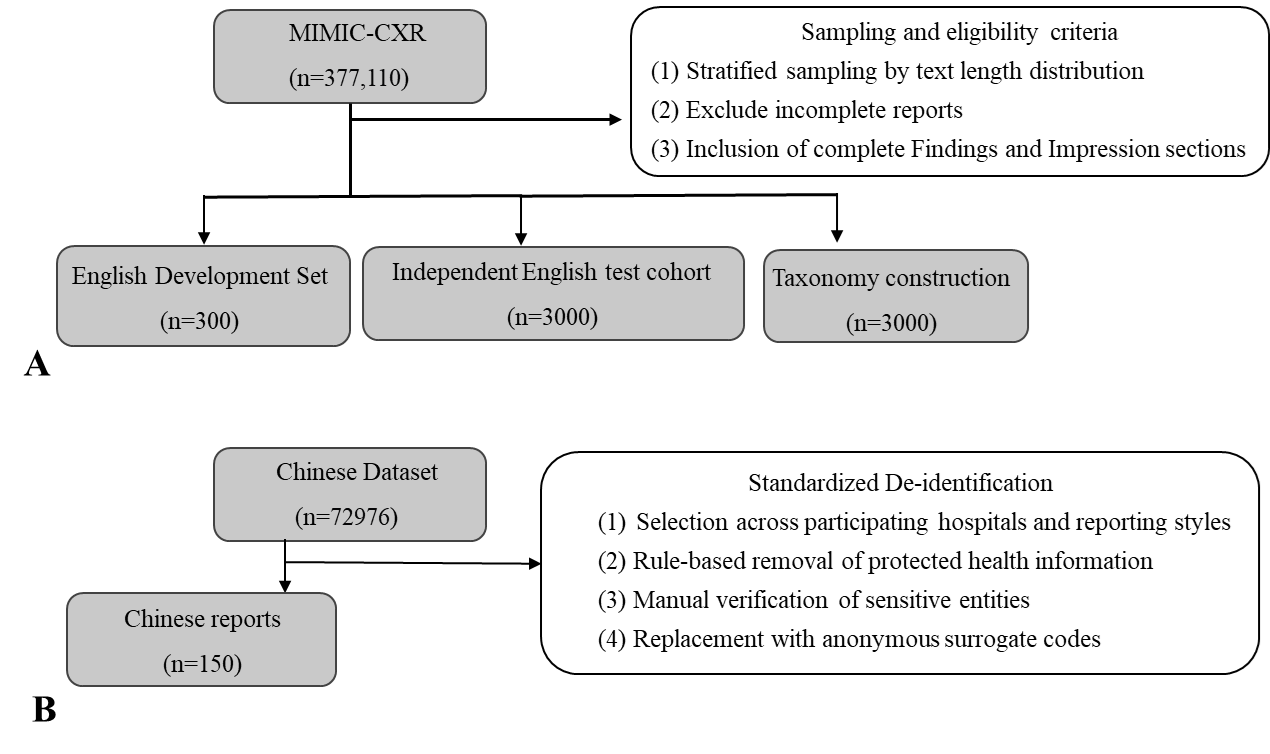}
  \caption{Flowchart of dataset construction and cohort allocation. Reports from MIMIC-CXR were screened and divided into three non-overlapping MIMIC-CXR-EN cohorts: a 3,000-report cohort for taxonomy construction, a 300-report development cohort for prompt optimization and radiologist reference-standard-based evaluation, and an independent 3,000-report test cohort for downstream benchmarking of report generation models. An additional 150-report ChestX-CN validation cohort was used for external validation.}
  \label{fig:dataset_flow}
\end{figure}

\subsection{MIMIC-CXR-EN Development Cohort}
On the 300-report MIMIC-CXR-EN development cohort, Qwen3-14B achieved the highest accuracy (Table~2), with $\geq 95\%$ accuracy on 17 of the 21 labels. This included 99.33\% (95\% CI 0.9761--0.9992) for Fracture, 99.00\% for Pleural Effusion, and 99.00\% for Cavity and Cyst. Qwen-Plus reached equivalent performance on Cardiomegaly and Consolidation. In contrast, GPT-4o-mini and DeepSeek-R1 achieved only 82.33\% and 86.67\%, respectively, on Enlarged Cardiomediastinum (compared with 95.67\% for Qwen3-14B).

Qwen3-14B showed higher performance, consistent with the fact that the prompt was optimized for this model. The other models showed clear performance deficits, most notably the $> 50$ percentage point difference for GPT-3.5-turbo on Cardiomegaly. These results highlight the critical role of domain-specific prompt engineering in medical natural language processing.

\begin{table}[!htbp]
  \centering
  \caption{\textbf{Per-label accuracy of LLM-based finding extraction on the 300-report MIMIC-CXR-EN development cohort.}}
  \label{tab:mimic_dev}
  \scriptsize   
  \setlength{\tabcolsep}{6pt}

  \begin{tabular}{l*{5}{c}}
    \toprule
    \centering
    Column Name & Qwen3-14B & Qwen-Plus & GPT-3.5-turbo & GPT-4o-mini & DeepSeek-R1 \\
    \midrule
    Atelectasis & 0.987 [0.966, 0.996] & 0.980 [0.957, 0.993] & 0.787 [0.736, 0.832] & 0.980 [0.957, 0.993] & 0.973 [0.948, 0.988] \\
    Cardiomegaly & 0.980 [0.957, 0.993] & 0.987 [0.966, 0.996] & 0.477 [0.419, 0.535] & 0.987 [0.966, 0.996] & 0.967 [0.940, 0.984] \\
    Consolidation & 0.973 [0.948, 0.988] & 0.983 [0.962, 0.995] & 0.787 [0.736, 0.832] & 0.977 [0.953, 0.991] & 0.990 [0.971, 0.998]\\
    Edema & 0.980 [0.957, 0.993] & 0.977 [0.953, 0.991] & 0.623 [0.566, 0.678] & 0.980 [0.957, 0.993] & 0.983 [0.962, 0.995]\\
    Enlarged Cardiomediastinum & 0.957 [0.927, 0.977] & 0.980 [0.957, 0.993] & 0.780 [0.729, 0.826] & 0.823 [0.775, 0.865] & 0.867 [0.823, 0.903]\\
    Fracture & 0.993 [0.976, 0.999] & 0.990 [0.971, 0.998] & 0.947 [0.915, 0.969] & 0.990 [0.971, 0.998] & 0.990 [0.971, 0.998]\\
    Lung Lesion & 0.963 [0.935, 0.982] & 0.970 [0.944, 0.986] & 0.663 [0.607, 0.717] & 0.880 [0.838, 0.915] & 0.907 [0.868, 0.937]\\
    No Finding & 0.930 [0.895, 0.956] & 0.963 [0.935, 0.982] & 0.783 [0.732, 0.829] & 0.940 [0.907, 0.964] & 0.970 [0.944, 0.986]\\
    Pleural Effusion & 0.990 [0.971, 0.998] & 0.987 [0.966, 0.996] & 0.847 [0.801, 0.886] & 0.970 [0.944, 0.986] & 0.967 [0.940, 0.984]\\
    Pleural Other & 0.967 [0.940, 0.984] & 0.990 [0.971, 0.998] & 0.567 [0.509, 0.624] & 0.900 [0.860, 0.932] & 0.987 [0.966, 0.996]\\
    Lung Opacity & 0.943 [0.911, 0.967] & 0.950 [0.919, 0.972] & 0.720 [0.666, 0.770] & 0.937 [0.903, 0.961] & 0.940 [0.907, 0.964]\\
    Pneumonia & 0.967 [0.940, 0.984] & 0.957 [0.927, 0.977] & 0.773 [0.722, 0.820] & 0.957 [0.927, 0.977] & 0.947 [0.915, 0.969]\\
    Pneumothorax & 0.983 [0.962, 0.995] & 1.000 [0.988, 1.000] & 0.943 [0.911, 0.967] & 1.000 [0.988, 1.000] & 1.000 [0.988, 1.000]\\
    Support Devices & 0.920 [0.883, 0.948] & 0.953 [0.923, 0.974] & 0.797 [0.747, 0.841] & 0.933 [0.899, 0.959] & 0.957 [0.927, 0.977]\\
    Emphysema & 0.993 [0.976, 0.999] & 0.983 [0.962, 0.995] & 0.967 [0.940, 0.984] & 0.990 [0.971, 0.998] & 0.993 [0.976, 0.999]\\
    Interstitial Lung Disease & 0.997 [0.982, 1.000] & 0.990 [0.971, 0.998] & 0.913 [0.876, 0.943] & 0.997 [0.982, 1.000] & 0.977 [0.953, 0.991]\\
    Calcification (Lung/Mediastinal) & 0.980 [0.957, 0.993] & 0.990 [0.971, 0.998] & 0.843 [0.797, 0.883] & 0.990 [0.971, 0.998] & 0.993 [0.976, 0.999]\\
    Trachea and Bronchus & 0.953 [0.923, 0.974] & 0.977 [0.953, 0.991] & 0.690 [0.634, 0.742] & 0.977 [0.953, 0.991] & 0.977 [0.953, 0.991]\\
    Cavity and Cyst & 0.990 [0.971, 0.998] & 0.993 [0.976, 0.999] & 0.803 [0.754, 0.847] & 0.993 [0.976, 0.999] & 0.990 [0.971, 0.998]\\
    Mediastinal Other & 0.947 [0.915, 0.969] & 0.950 [0.919, 0.972] & 0.873 [0.830, 0.909] & 0.960 [0.931, 0.979] & 0.947 [0.915, 0.969]\\
    Pulmonary Vascular Abnormal & 0.960 [0.931, 0.979] & 0.957 [0.927, 0.977] & 0.860 [0.816, 0.897] & 0.963 [0.935, 0.982] & 0.917 [0.879, 0.945]\\
    
    \bottomrule
    \addlinespace[4pt]
    \multicolumn{6}{@{}p{\linewidth}@{}}{\footnotesize \raggedright \textit{Note.} --- Values are accuracy rates with 95\% confidence intervals in brackets. Accuracy was calculated independently for each finding label. CI, confidence interval.} \\
  \end{tabular}
\end{table}

\subsection{ChestX-CN Validation Cohort}
All models were evaluated on the ChestX-CN validation cohort using the same few-shot prompt template that had been optimized and frozen on the MIMIC-CXR-EN reports, with no language-specific modifications applied. This cross-lingual evaluation was designed to test the inherent generalization capability of the prompt engineering framework.
On the ChestX-CN validation cohort, the same frozen few-shot prompt---optimized solely on the MIMIC-CXR-EN reports---was applied directly, without any language-specific modifications or additional exemplars. Qwen3-14B demonstrated robust zero-shot cross-lingual generalization, achieving $\geq 95\%$ accuracy on 16 of the 21 labels (Table~3). This included 100\% accuracy on several labels, such as Edema, Enlarged Cardiomediastinum and Pulmonary Vascular Abnormality.
Compared with the 300-report MIMIC-CXR-EN development cohort (where $\geq 95\%$ accuracy was reached on 17 of 21 labels), performance on the ChestX-CN cohort showed modest reductions for certain common abnormalities (for example, Lung opacity 89.3\%; Pneumonia 89.3\%), most likely attributable to differences in terminology and narrative style between the languages. Nevertheless, recall for rare abnormalities remained 100\%, and overall agreement with the radiologist-derived reference standard remained high for the majority of labels.
In contrast, the other models exhibited more marked performance drops on the ChestX-CN validation cohort (for example, GPT-3.5-turbo showed substantial declines on several key labels). These findings underscore the importance of clinician-guided prompt engineering for achieving cross-lingual robustness.

\begin{table}[!htbp]
  \centering
  \caption{\textbf{Per-label accuracy of LLM-based finding extraction on the 150-report ChestX-CN validation cohort.}}
  \label{tab:chestxcn}
  \scriptsize   
  \setlength{\tabcolsep}{6pt}

  \begin{tabular}{l*{5}{c}}
    \toprule
    \centering
    Column Name & Qwen3-14B & Qwen-Plus & GPT-3.5-turbo & GPT-4o-mini & DeepSeek-R1 \\
    \midrule
    Atelectasis & 0.993 [0.976, 0.999] & 0.920 [0.883, 0.948] & 0.273 [0.224, 0.328] & 0.940 [0.907, 0.964] & 0.987 [0.966, 0.996]\\
    Cardiomegaly & 0.980 [0.957, 0.993] & 0.993 [0.976, 0.999] & 0.153 [0.115, 0.199] & 0.987 [0.966, 0.996] & 0.993 [0.976, 0.999] \\
    Consolidation & 0.987 [0.966, 0.996] & 0.960 [0.931, 0.979] & 0.460 [0.403, 0.518] & 0.960 [0.931, 0.979] & 0.980 [0.957, 0.993]\\
    Edema & 1.000 [0.988, 1.000] & 1.000 [0.988, 1.000] & 0.267 [0.218, 0.321] & 1.000 [0.988, 1.000] & 1.000 [0.988, 1.000]\\
    Enlarged Cardiomediastinum & 1.000 [0.988, 1.000] & 0.993 [0.976, 0.999] & 0.700 [0.645, 0.751] & 0.993 [0.976, 0.999] & 0.993 [0.976, 0.999]\\
    Fracture & 0.993 [0.976, 0.999] & 0.973 [0.948, 0.988] & 0.947 [0.915, 0.969] & 1.000 [0.988, 1.000] & 0.980 [0.957, 0.993]\\
    Lung Lesion & 0.920 [0.883, 0.948] & 0.907 [0.868, 0.937] & 0.687 [0.631, 0.739] & 0.953 [0.923, 0.974] & 0.780 [0.729, 0.826]\\
    No Finding & 0.947 [0.915, 0.969] & 0.913 [0.876, 0.943] & 0.900 [0.860, 0.932] & 0.947 [0.915, 0.969] & 0.927 [0.891, 0.954]\\
    Pleural Effusion & 0.960 [0.931, 0.979] & 0.953 [0.923, 0.974] & 0.453 [0.396, 0.512] & 0.967 [0.940, 0.984] & 0.960 [0.931, 0.979]\\
    Pleural Other & 0.907 [0.868, 0.937] & 0.980 [0.957, 0.993] & 0.293 [0.242, 0.348] & 0.693 [0.638, 0.745] & 0.973 [0.948, 0.988]\\
    Lung Opacity & 0.893 [0.853, 0.926] & 0.893 [0.853, 0.926] & 0.707 [0.652, 0.758] & 0.760 [0.708, 0.807] & 0.693 [0.638, 0.745]\\
    Pneumonia & 0.893 [0.853, 0.926] & 0.900 [0.860, 0.932] & 0.540 [0.482, 0.597] & 0.827 [0.779, 0.868] & 0.827 [0.779, 0.868]\\
    Pneumothorax & 0.973 [0.948, 0.988] & 0.960 [0.931, 0.979] & 0.640 [0.583, 0.694] & 0.960 [0.931, 0.979] & 0.967 [0.940, 0.984]\\
    Support Devices & 0.967 [0.940, 0.984] & 0.940 [0.907, 0.964] & 0.820 [0.772, 0.862] & 0.907 [0.868, 0.937] & 0.953 [0.923, 0.974]\\
    Emphysema & 0.993 [0.976, 0.999] & 1.000 [0.988, 1.000] & 0.767 [0.715, 0.813] & 0.993 [0.976, 0.999] & 1.000 [0.988, 1.000]\\
    Interstitial Lung Disease & 1.000 [0.988, 1.000] & 0.980 [0.957, 0.993] & 0.393 [0.338, 0.451] & 0.973 [0.948, 0.988] & 0.993 [0.976, 0.999]\\
    Calcification (Lung/Mediastinal) & 0.973 [0.948, 0.988] & 0.967 [0.940, 0.984] & 0.520 [0.462, 0.578] & 0.967 [0.940, 0.984] & 0.960 [0.931, 0.979]\\
    Trachea and Bronchus & 0.987 [0.966, 0.996] & 0.973 [0.948, 0.988] & 0.327 [0.274, 0.383] & 1.000 [0.988, 1.000] & 0.980 [0.957, 0.993]\\
    Cavity and Cyst & 1.000 [0.988, 1.000] & 0.900 [0.860, 0.932] & 0.440 [0.383, 0.498] & 0.987 [0.966, 0.996] & 0.967 [0.940, 0.984]\\
    Mediastinal Other & 0.993 [0.976, 0.999] & 0.993 [0.976, 0.999] & 0.600 [0.542, 0.656] & 0.980 [0.957, 0.993] & 0.980 [0.957, 0.993]\\
    Pulmonary Vascular Abnormal & 1.000 [0.988, 1.000] & 1.000 [0.988, 1.000] & 0.667 [0.610, 0.720] & 1.000 [0.988, 1.000] & 1.000 [0.988, 1.000]\\

    \bottomrule
    \addlinespace[4pt]
    \multicolumn{6}{@{}p{\linewidth}@{}}{\footnotesize \raggedright \textit{Note.} --- Values are accuracy rates with 95\% confidence intervals in brackets. The ChestX-CN validation cohort was not used during prompt optimization. CI, confidence interval.} \\
  \end{tabular}
\end{table}

\subsection{Performance After Few-Shot Prompt Optimization}
Targeted prompt optimization substantially improved the performance of Qwen3-14B on the MIMIC-CXR-EN report extraction task, as shown in Table~4. The macro-averaged F1 score increased from 0.753 to 0.956, representing an improvement of 20.3 percentage points, and exceeded the CheXbert reference standard by 15.7 percentage points, while overall accuracy remained high and increased from 96.9\% to 98.8\%. The largest gains were observed in low-prevalence categories, with F1 improvements of 79.8 percentage points for Trachea and Bronchus, 57.6 percentage points for Mediastinal Other, 55.6 percentage points for Pneumothorax, 62.5 percentage points for Pleural Other, and 33.3 percentage points for Fracture. Recall for Fracture increased from 50\% to 100\%. Qwen3-14B outperformed CheXbert on all 14 directly comparable labels, with the largest gains observed for Lung Lesions at 66.4 percentage points and Pleural Other at 53.4 percentage points. After optimization, several entities, including Fracture, Pneumothorax, and Cavity and Cyst, achieved perfect F1 scores of 1.000. These findings support the use of macro-averaged F1 score as the primary metric in imbalanced medical datasets and suggest that few-shot prompt engineering can reduce performance disparities across underrepresented abnormalities.

\begin{table}[!htbp]
  \centering
  \caption{\textbf{Per-label performance of Qwen3-14B before and after few-shot prompt optimization on the 300-report MIMIC-CXR-EN development cohort.}}
  \label{tab:per-label-accuracy}
  
  \footnotesize
  \setlength{\tabcolsep}{4pt}
  
  \begin{tabular}{l *{12}{c}}
    \toprule
    
    % grouped header row
    \multirow{2}{*}{Column Name} & 
    \multicolumn{2}{c}{Accuracy} & 
    \multicolumn{2}{c}{Precision} & 
    \multicolumn{2}{c}{Recall} & 
    \multicolumn{2}{c}{F1 score} & 
    $\Delta$ F1 & CheXbert & $\Delta$ F1 \\
    \cmidrule(lr){2-3} \cmidrule(lr){4-5} \cmidrule(lr){6-7} \cmidrule(lr){8-9} \cmidrule(lr){10-10} \cmidrule(lr){11-11} \cmidrule(lr){12-12}
    
    % second header row
    
    \midrule
    Atelectasis & 0.987 & 0.987 & 0.957 & 0.957 & 1.000 & 1.000 & 0.978 & 0.978 & 0.000 & 0.940 & 0.013 \\
    Cardiomegaly & 0.980 & 0.987 & 0.987 & 0.986 & 0.937 & 0.960 & 0.961 & 0.973 & 0.013 & 0.815 & 0.159\\
    Consolidation & 0.973 & 0.993 & 0.900 & 0.950 & 0.750 & 0.950 & 0.818 & 0.950 & 0.087 & 0.877 & 0.028\\
    Edema & 0.980 & 0.990 & 0.939 & 0.919 & 0.886 & 1.000 & 0.912 & 0.958 & 0.018 & 0.881 & 0.049\\
    Enlarged Cardiomediastinum & 0.957 & 0.970 & 0.882 & 0.904 & 0.923 & 0.971 & 0.902 & 0.936 & 0.029 & 0.713 & 0.218\\
    Fracture & 0.993 & 1.000 & 1.000 & 1.000 & 0.500 & 1.000 & 0.667 & 1.000 & 0.333 & 0.791 & 0.209\\
    Lung Lesion & 0.963 & 0.983 & 0.714 & 0.862 & 0.870 & 0.962 & 0.784 & 0.909 & 0.159 & 0.664 & 0.279\\
    No Finding & 0.930 & 0.970 & 0.797 & 0.941 & 0.887 & 0.928 & 0.840 & 0.934 & 0.092 & 0.640 & 0.292\\
    Pleural Effusion & 0.990 & 0.993 & 0.963 & 1.000 & 1.000 & 0.976 & 0.981 & 0.988 & 0.001 & 0.919 & 0.063\\
    Pleural Other & 0.967 & 0.997 & 0.500 & 1.000 & 0.300 & 0.833 & 0.375 & 0.909 & 0.625 & 0.534 & 0.466\\
    Lung Opacity & 0.943 & 0.950 & 0.938 & 0.883 & 0.862 & 0.954 & 0.898 & 0.917 & 0.012 & 0.741 & 0.169\\
    Pneumonia & 0.967 & 0.990 & 0.900 & 0.969 & 0.794 & 0.939 & 0.844 & 0.954 & 0.097 & 0.835 & 0.106\\
    Pneumothorax & 0.983 & 1.000 & 1.000 & 1.000 & 0.286 & 1.000 & 0.444 & 1.000 & 0.556 & 0.928 & 0.072\\
    Support Devices & 0.920 & 0.987 & 0.860 & 0.975 & 0.925 & 0.992 & 0.891 & 0.983 & 0.047 & 0.888 & 0.050\\
    Emphysema & 0.993 & 0.997 & 0.900 & 0.909 & 0.900 & 1.000 & 0.900 & 0.952 & 0.100 & - & -\\
    Interstitial Lung Disease & 0.997 & 0.997 & 1.000 & 1.000 & 0.941 & 1.000 & 0.970 & 1.000 & 0.030 & - & -\\
    Calcification (Lung/Mediastinal) & 0.980 & 0.997 & 0.688 & 0.947 & 0.917 & 1.000 & 0.786 & 0.973 & 0.123 & - & -\\
    Trachea and Bronchus & 0.953 & 0.997 & 0.125 & 0.875 & 0.125 & 1.000 & 0.125 & 0.933 & 0.798 & - & -\\
    Cavity and Cyst & 0.990 & 1.000 & 0.500 & 1.000 & 0.667 & 1.000 & 0.571 & 1.000 & 0.429 & - & -\\
    Mediastinal Other & 0.947 & 0.990 & 0.222 & 0.944 & 0.667 & 0.895 & 0.333 & 0.919 & 0.576 & - & -\\
    Pulmonary Vascular Abnormal & 0.960 & 0.990 & 0.906 & 0.971 & 0.763 & 0.943 & 0.829 & 0.957 & 0.083 & - & -\\
    Average & 0.969 & 0.988 & 0.794 & 0.950 & 0.757 & 0.964 & 0.753 & 0.956 & 0.203 & 0.798 & 0.157\\

  \bottomrule
  \addlinespace[4pt]
  \multicolumn{13}{@{}p{\linewidth}@{}}{\small \raggedright \textit{Note.} --- Values rounded to three decimal places. Bold indicates best result per column. $\Delta$ F1 = post -- pre F1-score. CheXbert F1 and improvement columns are shown only for the 14 directly comparable labels.} \\
  \end{tabular}
\end{table}

\subsection{Evaluation of generated reports with Ran Score}
All models produced reports that were clinically coherent to varying degrees, but quantitative performance differed markedly, as shown in Table~5 and Figs.~3 and 4. We applied Ran Score, the proposed automated evaluation metric defined as the macro-averaged F1 score between standardized finding labels extracted from generated reports and those extracted from the corresponding original reference reports using the optimized Human--LLM collaborative framework. Ran Score ranked LLM-RG4 highest overall. Among the micro-averaged metrics, LLM-RG4 achieved the highest F1 score of 0.498 and the highest recall of 0.562, whereas PromptMRG achieved the highest precision at 0.568 and the highest accuracy at 0.881. R2GenGPT ranked next with a micro-averaged F1 score of 0.462, followed by MedKit with 0.382, while XrayGPT showed the lowest performance with a micro-averaged F1 score of 0.127. Macro-averaged metrics were consistently lower across all models, underscoring the persistent difficulty of detecting rare abnormalities. LLM-RG4 again performed best with a macro-averaged F1 score of 0.333, followed by Libra at 0.285 and R2GenGPT at 0.247; MedKit scored 0.164, and XrayGPT again performed worst at 0.045. Qualitative review by senior thoracic radiologists of 50 randomly selected reports per model was consistent with the quantitative ranking. LLM-RG4 and R2GenGPT were judged the most clinically appropriate and concise, followed by Libra and RGRG. MedKit was considered acceptable but less refined, whereas XrayGPT was considered clinically inapplicable, PromptMRG overly colloquial, and RGRG and Libra occasionally verbose or poorly organized. The broad agreement between automated scores and expert qualitative judgment supports the clinical relevance of the evaluation framework.

\begin{table}[!htbp]
  \centering
  \caption{\textbf{Performance comparison of radiology report generation models on the MIMIC-CXR-EN test cohort.}}
  \footnotesize
  \setlength{\tabcolsep}{5pt}
  
  \begin{tabular}{l *{8}{c}}
    \toprule
    
    % grouped header row
    \multirow{2}{*}{} & 
    \multicolumn{4}{c}{Micro-Averaged Performance} & 
    \multicolumn{3}{c}{Macro-Averaged Performance} \\
    \cmidrule(lr){2-5} \cmidrule(lr){6-8}
    & Accuracy & Precision & Recall & F1 score & Accuracy & Recall & F1 score \\
    % second header row
    \midrule
    RGRG & 0.874 & 0.391 & 0.512 & 0.443 & 0.216 & 0.306 & 0.220\\
    XrayGPT & 0.839 & 0.091 & 0.209 & 0.127 & 0.056 & 0.117 & 0.045\\
    R2GenGPT & 0.877 & 0.411 & 0.528 & 0.462 & 0.230 & 0.388 & 0.247\\
    PromptMRG & 0.881 & 0.327 & 0.568 & 0.415 & 0.172 & 0.342 & 0.178\\
    LLM-RG4 & 0.854 & 0.562 & 0.447 & 0.498 & 0.390 & 0.353 & 0.333\\
    Libra & 0.861 & 0.451 & 0.457 & 0.454 & 0.286 & 0.296 & 0.285\\
    MedKit & 0.868 & 0.317 & 0.481 & 0.382 & 0.165 & 0.235 & 0.164\\
    \bottomrule
  \end{tabular}

  \vspace{4pt}
  \begin{minipage}{0.98\linewidth}
    \footnotesize\raggedright \textit{Note.} --- Values are micro-averaged and macro-averaged precision, recall, and F1 scores. Micro-averaged metrics weight finding labels by prevalence, whereas macro-averaged metrics treat all labels equally.
  \end{minipage}
\end{table}

\begin{figure}[!htbp]               
  \centering
  \includegraphics[width=0.85\textwidth]{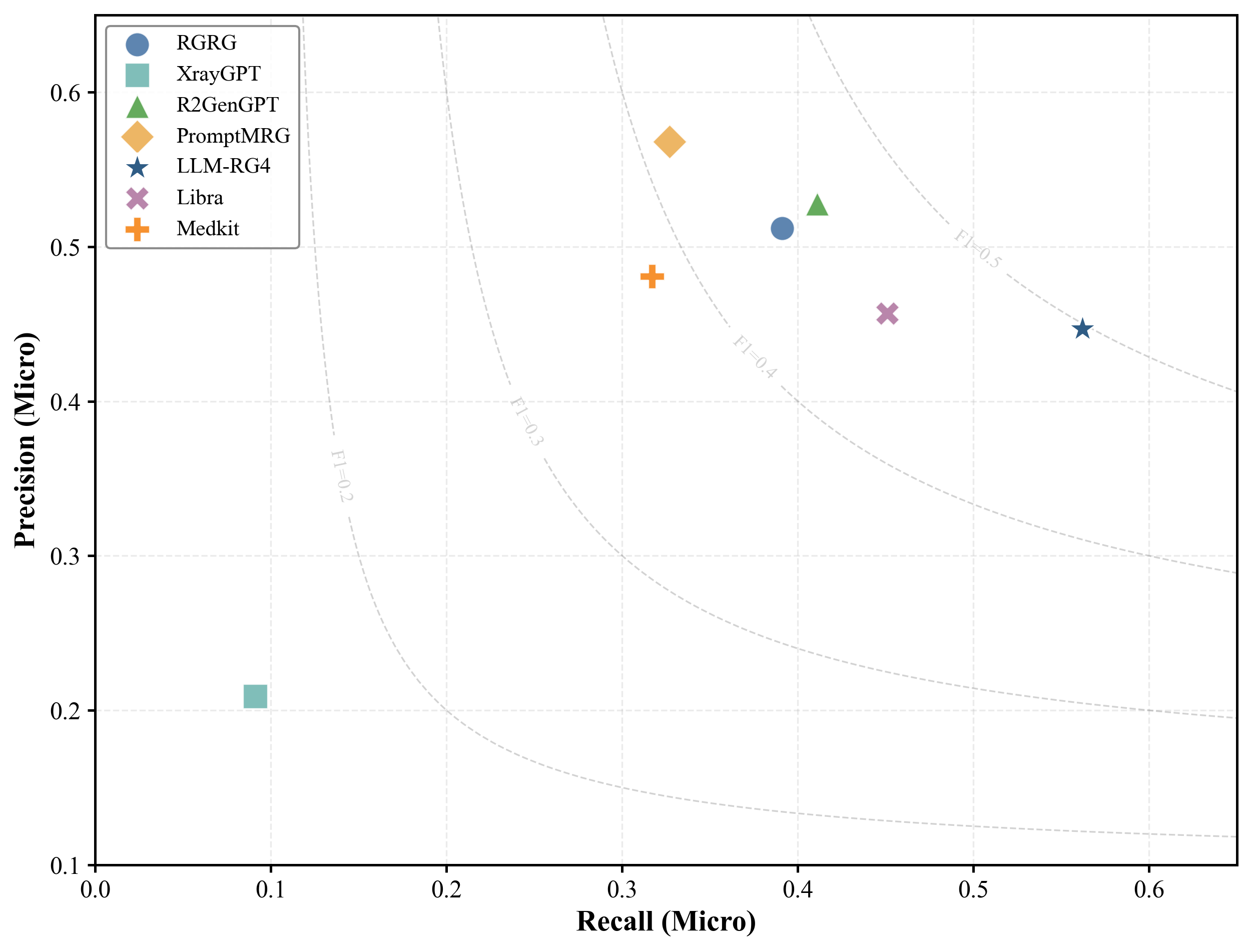}
  \caption{Micro-averaged performance of radiology report generation models on the MIMIC-CXR-EN test cohort. Micro-averaged precision, recall and F1 score were calculated across all finding labels, with each label weighted according to its prevalence in the dataset. Model predictions were evaluated using the optimized Qwen3-14B labeler.}
  \label{fig:micro_models}
\end{figure}

\begin{figure}[!htbp]               
  \centering
  \includegraphics[width=0.85\textwidth]{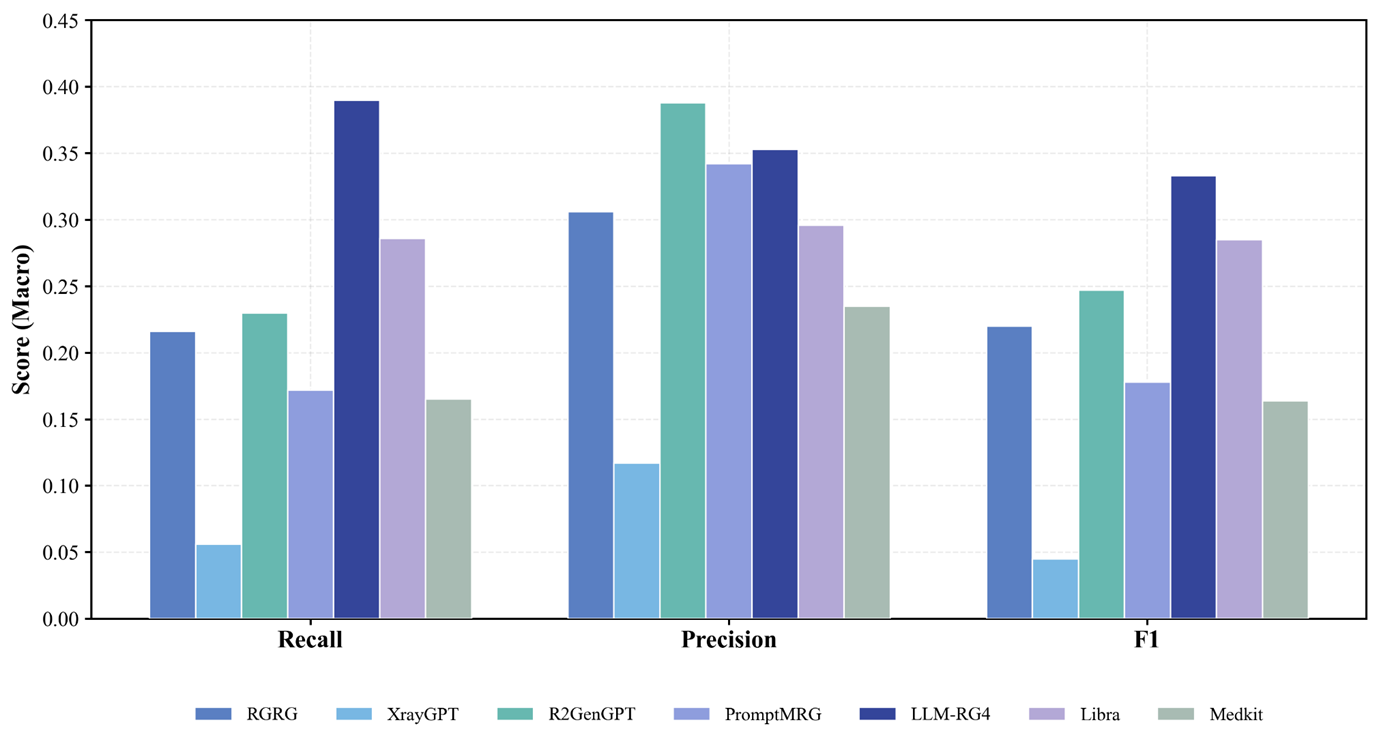}
  \caption{Macro-averaged performance of radiology report generation models across finding categories. Macro-averaged precision, recall and F1 score were calculated by treating all finding labels equally. This approach emphasizes model performance on low-prevalence and rare findings.}
  \label{fig:macro_models}
\end{figure}

\section{Discussion}

In this study, targeted clinician-guided prompt optimization enabled high agreement with the radiologist-derived reference standard for multi-label finding extraction from chest X-ray reports\cite{papineni2002,lin2004,vedantam2015}.Using a standardized 21-label taxonomy, the optimized Qwen3-14B framework achieved a macro-averaged F1 score of 0.956, with all pre-post improvements reaching statistical significance. Ran Score served as the primary evaluation metric and was defined as the macro-averaged F1 score derived from finding-level agreement between generated reports and their corresponding original reference reports. By assigning equal weight to each label, it is more sensitive to clinically important omissions in low-prevalence abnormalities and provides a clinically grounded measure of report fidelity\cite{yu2023,ma2024,decideai2021}.

The most substantial gains were observed for low-prevalence findings, which remain difficult for both rule-based systems and general-purpose language models. After optimization, the macro-averaged F1 score increased from 0.753 to 0.956 and exceeded the CheXbert benchmark by 15.7 percentage points on comparable labels. The largest improvements were seen for Pneumothorax, Pleural Other, Trachea and Bronchus, Mediastinal Other, and Fracture. Recall for Fracture increased from 50\% to 100\%, and several labels, including Fracture, Pneumothorax, and Cavity and Cyst, achieved F1 scores of 1.000. These findings support the use of macro-averaged evaluation, as implemented in Ran Score, for assessing performance on clinically important but underrepresented abnormalities.

The observed improvements were associated with iterative correction of specific failure modes through clinician-guided prompt refinement. For Pneumothorax, reports describing indirect evidence, such as a visceral pleural line with associated lung collapse, were often misclassified before optimization because the finding was not stated explicitly. After targeted positive and negative exemplars were introduced, the prompt more consistently identified such reports in agreement with radiologist interpretation. A similar pattern was observed for fracture labeling, in which descriptions such as healed fracture, old fracture, compression deformity, or stable findings on prior imaging were more reliably recognized after refinement. These examples suggest that clinician-guided prompt engineering can improve handling of synonym variation, negation, and contextual ambiguity without retraining the underlying model\cite{zhao2024,hager2024}.

Image-to-report generation systems showed limited performance under this evaluation framework. Their quantitative results were broadly consistent with qualitative assessments by senior thoracic radiologists, who judged many generated reports to be clinically unreliable. Although LLM-RG4 achieved the highest scores among the tested models, performance remained limited by incomplete detection of rare abnormalities and susceptibility to hallucinated content. Higher-scoring models were generally rated as more concise and clinically appropriate, whereas lower-performing models produced more verbose or less reliable reports. These findings suggest that conventional generation benchmarks do not fully capture diagnostic utility, whereas Ran Score provides a more clinically grounded assessment by emphasizing finding-level agreement, particularly for rare findings\cite{adams2023,sun2023}.

Several limitations should be noted. First, the study was retrospective and restricted to chest X-ray reports, and further validation across imaging modalities and institutions will be needed before clinical use\cite{vandesande2024}. Second, prompt optimization and reference-standard-based performance evaluation were both conducted on the same 300-report MIMIC-CXR-EN development cohort, which may overestimate extraction performance relative to evaluation on a fully independent annotated test set. Third, uncertain or equivocal findings were excluded, so the present framework does not fully represent the degree of diagnostic ambiguity encountered in routine practice. Fourth, because Ran Score relies on the same optimized extraction framework to derive standardized finding labels from both generated reports and their corresponding original reference reports, the metric may retain extractor-specific bias. Fifth, standard evaluation metrics may favor long or templated reports even when such reports are not considered clinically useful. Finally, the framework depends on iterative radiologist feedback, which may limit scalability in settings with limited specialist availability\cite{white2022,gundogdu2021}.

Despite these limitations, a major practical value of this work lies in its implications for chest X-ray report generation on MIMIC-CXR. Many existing report generation studies rely on CheXbert-derived labels for supervision or automated evaluation, yet the limited scope of the CheXbert label space may reduce sensitivity to clinically important findings, particularly low-prevalence or more fine-grained abnormalities. Because our framework produces broader and more clinically aligned multi-label annotations across the full MIMIC-CXR corpus, it may support stronger supervision, more sensitive evaluation, and improved report fidelity in future image-to-report models. To support reproducibility and further development in radiology AI, we publicly release the complete multi-label annotations for the MIMIC-CXR dataset together with the associated code, prompts, and guidelines. Overall, these findings suggest that radiologist-guided prompt optimization offers a practical route to more clinically reliable language-model evaluation.

\section*{Acknowledgments}
This study was supported bythe National Natural Science Foundation of China (grant no. 62306037, 62025104); the National Key Clinical Specialty Construction Project, China (grant no. 2024-QTL-001);and Diseases-National Science and Technology Major Project, China (grant nos. 2024ZD0529500 and 2024ZD0529503).

\end{document}